\documentclass{egpubl}

\WsSubmission    
 \electronicVersion 

\ifpdf \usepackage[pdftex]{graphicx} \pdfcompresslevel=9
\else \usepackage[dvips]{graphicx} \fi

\PrintedOrElectronic

\usepackage{t1enc,dfadobe}

\usepackage{egweblnk}
\usepackage{cite}
\usepackage{amsmath}

\title[A Guided Spatial Transformer Network for Histology Cell Differentiation]%
      {A Guided Spatial Transformer Network for Histology Cell Differentiation}

\author[M. Aubreville, M. Krappmann, C. Bertram, R. Klopfleisch \& A. Maier]
	{Marc Aubreville$^{1}$,
	Maximilian Krappmann$^{1}$,
	Christof Bertram$^{2}$,
	Robert Klopfleisch$^{2}$ and
	Andreas Maier$^{1}$ 
	\\
         $^1$Pattern Recognition Lab, Friedrich-Alexander-Universit{\"a}t Erlangen-N{\"u}rnberg, Germany \\
         $^2$Institute of Veterinary Pathology, Free University Berlin, Germany
}	

\ConfYear{2017}
\ConfName{}
\begin{document}

\maketitle

\begin{abstract}

Identification and counting of cells and mitotic figures is a standard task in diagnostic histopathology. Due to the large overall cell count on histological slides and the potential sparse prevalence of some relevant cell types or mitotic figures, retrieving annotation data for sufficient statistics is a tedious task and prone to a significant error in assessment. Automatic classification and segmentation is a classic task in digital pathology, yet it is not solved to a sufficient degree. 

We present a novel approach for cell and mitotic figure classification, based on a deep convolutional network with an incorporated Spatial Transformer Network. The network was trained on a novel data set with ten thousand mitotic figures, about ten times more than previous data sets. The algorithm is able to derive the cell class (mitotic tumor cells, non-mitotic tumor cells and granulocytes) and their position within an image. The mean accuracy of the algorithm in a five-fold cross-validation is 91.45\,\%. 

In our view, the approach is a promising step into the direction of a more objective and accurate, semi-automatized mitosis counting supporting the pathologist.


\end{abstract}

\section{Introduction}

The assessment of cell types in histology slides is a standard task in pathology. Especially in tumor diagnostics, determining the relative amount of mitotic figures, a marker for tumor proliferation and aggressiveness, is another important task for the diagnostic pathologist \cite{Roux:2013kn}.

However, evaluation of the complete slide for mitotic figures is usually too time consuming in routine diagnostics. Therefore it is suggested that only 10 high power fields (an area of assumed equal size used for statistic comparison), presumed to contain the highest density of mitoses, are subjectively chosen by the pathologist. The area of these fields is, however, not well-defined, as it depends on the optical properties of the microscope and which may vary significantly in their content of mitotic figures \cite{Meuten:2016jh}. The final count thus strongly depends on the randomly but not necessarily representatively selected high power fields thus the resulting mitotic count is usually observer-dependent \cite{Veta:2014bi}. In addition, mitotic figures may be very variable in their histologic phenotype, which may also lead to inter-observer variability between pathologists. 

The aim of this work is to develop a more objective and accurate, automatized approach to counting of mitotic figures by assisting pathologists in the selection of fields with the highest mitotic counts and with more constant parameters of mitotic figure identification. 
Detection and annotation of mitotic cells in histology slides is a well-known task in images processing, and subject of several challenges in recent years \cite{Veta:2014bi}\cite{Roux:2013kn}.

\begin{figure}[htb]
  \centering
  \includegraphics[width=.9\linewidth]{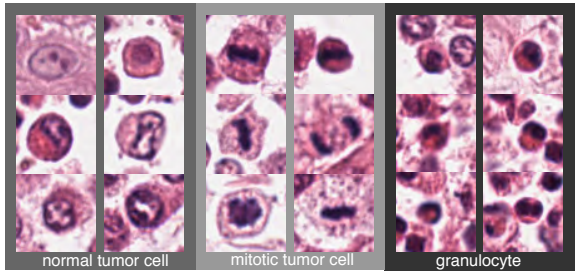}

  \caption{\label{fig:mitosis}
           Examples of cropped cells, slides stained with hematoxylin and eosin. }
\end{figure}

Mitosis comprises a number of different phases in the cell cycle (prophase, metaphase, anaphase, and telophase). In each phase, the nucleus is shaped differently. This means that the variance in images showing a mitotic cell is high (see figure \ref{fig:mitosis}). On top of that, there is also atypical mitosis, adding yet another factor of variance to the picture. However, publicly available databases for mitosis detection feature a rather low number of mitotic figures (e.g. the 2014 ICPR MITOS-ATYPIA-14 dataset with 873 images,  the 2012 ICPR dataset with 326 images\cite{Roux:2013kn}, or the AMIDA13 dataset with 1083 images\cite{Veta:2014bi}), especially for robust detection.

Automatic detection of mitotic figures has been widely performed using the classical machine learning workflow on textural, morphological and shape features (e.g. \cite{Sommer:2012wy}\cite{Irshad:2013tu}).Cire\c{s}an \textit{et al.} were the first to employ deep learning-based approaches for mitosis detection \cite{Ciresan:2013up}, yielding significant improvements over traditional approaches \cite{Veta:2014bi}. Yet, deep learning technologies suffer considerably from insufficient data amounts, as they have a large number of trainable parameters and, because of this, are likely to overfit the data.
Particularly in the field of mitosis detection, we assume that detection performance could be improved if the whole variance of mitotic processes can be captured in the networks, requesting for a substantial increase in training data for such networks.

\section{Related Work}

Typically, the process of object detection is parted into two sub-processes: Segmentation and classification. This setup is especially sensible for histology since the images represent a large amount of data and classification is usually the more complex process compared to segmentation. Sommer \textit{et al.} used pixel-wise classification for candidate retrieval and then object shape and texture features for mitotic cell classification \cite{Sommer:2012wy}. Irshad used active contour models for candidate selection and statistical and morphological features for classification \cite{Irshad:2013tu}. Those hand-crafted features have significant drawbacks, however: Given the often small data sets, automatic selection of features is prone to random correlation, while using higher-dimensional classification approaches on the complete set increases overfitting \cite{Leardi:1996ez}. Further, it is questionable, if those approaches can represent the variability in shape and texture of mitotic figures \cite{Chen:2016vv}.

Triggered by the ground-breaking initial works of Lecun \cite{LeCun:1998fv}, Convolutional Neural Networks (CNN) have spread widely in the use for various image classification tasks. CNN-based recognition algorithms have won all major image recognition challenges in recent years because of their ability to capture complex shapes and still remain sensitive to minor variations in the image. In the field of mitosis detection, CNN-based approaches have been used for classification \cite{Ciresan:2013up}, feature extraction \cite{Wang:2014ka} as well as  candidate generation \cite{Chen:2016vv}. Yet, CNNs, through their inherent ability to capture complex structures, are also prone to overfitting, a problem which is usually targeted by data augmentation and regularization strategies like dropout and other mechanisms or by means of transfer learning. Another regularization strategy is to constrain the capacity of the approach \cite{goodfellow2016deep} by reducing effectively the free parameters of the model. We aim to attempt this by splitting the problem into an attention task and a classification task. The general issue however, that the training data might be a non-representative sample of the classification task and thus parts of real-world data are not recognized because the data set does not generalize well, can be best targeted with a bigger training data set, as it was the base for this work.



\section{Material}

For this study, digital histopathological images were acquired using Aperio ScanScope (Leica Biosystems Imaging, Inc., USA) slide scanner hardware at a magnification of 400x. Candidate patches for three different cell types (mitotic cells, eosinophilic granulocytes and normal tumor cells) were annotated by an expert with profound knowledge on cell differentiation and classified by a trained pathologist. The cells were selected from histologic images of 12 different paraffin-embedded canine mast cell tumors, stained with standard hematoxylin and eosin (H\&E). In order to train a deep learning classifier with a sufficient amount of data, our emphasis was not on complete annotation of the slides but on finding enough candidates for the above-mentioned cell types within the image. 

More specifically, the emphasis was on finding mitotic cells. Commonly, in all major related works, the number of mitotic cells in the data set was proportional to the actual occurrence in the respective slides, as whole slides where annotated, resulting in a relatively low number of mitotic cells. On the contrary, we purposely selected a similar number of cells from each category to not assume any priors in distribution. 

We acknowledge that this procedure might add a certain bias in cell selection, and that our dataset might not be representative. However, this argument can also be made for the case where only a small number of mitotic figures is available. Further, because of the high inter-rater variability in mitosis expert classifications, we assume that an unbiased ground truth is hard to retrieve and a minor bias by image pre-selection can be tolerated. Finally, we do not target at finding all mitotic cells, but rather to guide the pathologist in finding a representative part of the slide and to thus reduce variability in expert grading.

In the data set, we have approx. 37,800 single annotations of cells of the three different types (about 10,400 granulocytes, 10,800 mitotic figures and 16,600 normal tumor cells). The majority of cells was rated by the pathologist to be normal tumor cells, however also a significant amount of mitotic cells and eosinophil granulocytes was annotated. 

\begin{figure}[htb]
  \centering
  \includegraphics[width=.8\linewidth]{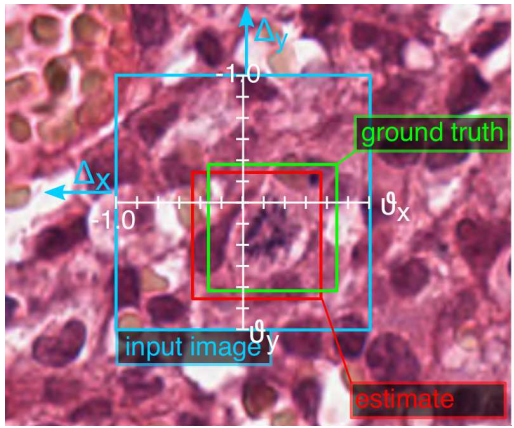}

  \caption{\label{fig:randomCropping}
           Image preprocessing. The offset $\Delta_x$, $\Delta_y$   is set randomly while keeping the cell within the image.  }
\end{figure}

\section{Methods}

Spatial Transformer Networks (STN), first described by Jaderberg et. al, provide a learnable method to focus the attention of a classification network on a specific subpart of the original image \cite{Jaderberg:J0WMXL0g}. To achieve this, parameters of an affine transformation matrix $\theta$ are regressed by the network, alongside with the optimization of the actual classifier.  

Spatial Transformer Networks were originally successfully employed on a distorted MNIST data set, where translation, scale, rotation and clutter were used to increase the difficulty for the detection task. The approach has shown to be able to -- without any prior knowledge about the actual transform that was applied beforehand -- increase accuracy of the classification network by focusing its attention to the area where the number was present and by compensating for the deformation \cite{Jaderberg:J0WMXL0g}. The approach can be used in a joined learning approach, where both the transform and the classification are learned end-to-end, something that could be described as a weakly supervised learning approach for the transformation. The optimization on the MNIST data set is, however, a much easier task than on real-world data. In a typical patch extracted from a histology slide, a lot of similar and valid objects may be contained in the image, and joined optimization suffers from local extrema in the gradient descend approach.

In this work, we aim to use STN as a method of not only directing the attention of a classification network to a sub-area of a larger image, and thus hopefully improving classification performance, but also as a segmentation approach to derive the information about where the respective cells are located.

We believe that Spatial Transformer Networks are an ideal candidate for this kind of task because they can be used to model two sources of natural variance into the machine learning process with a comparatively small overhead in complexity: Scaling and translation. Scaling is relevant in microscopy for two reasons: Firstly, the actual magnification of the microscope is dependent on the optical properties of the ocular, notably on the field number\cite{Meuten:2016jh}. Secondly, cells differ in size, dependent on their function and the species they originate from.  

\begin{figure}[htb]
  \centering
  \includegraphics[width=.99\linewidth]{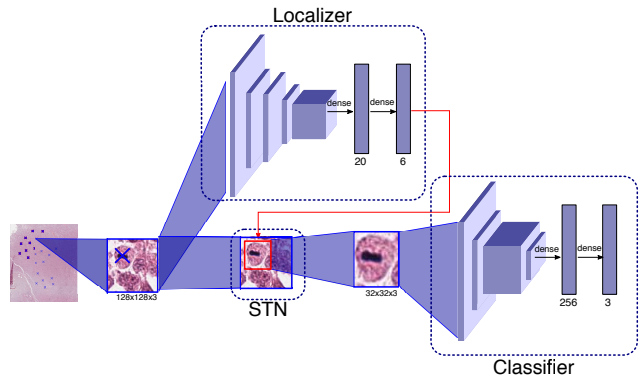}
  \caption{\label{fig:network}
           Overview of the network. The cell's position is estimated by the localizer, which regresses an affine transform applied on the original image, to feed the classifier with cell images.}
\end{figure}

\subsection{Image Preprocessing}

\label{chp:methods}
All images were cropped around the cell center in a first processing step. In a second processing step, we introduce a random translation $\Delta_x$, $\Delta_y$ to the origin area of the input image before cropping, so it is no longer centered around the cell, i.e. the cell can be anywhere on the image, with the restriction that the whole cell will be within the image (see figure \ref{fig:randomCropping}). From the introduced translation, we can derive a new ground truth transformation vector 

\begin{equation}
	\theta = \left[ \begin{array}{ccc}
		\vartheta_s & 0 & \vartheta_x \\
		0 & \vartheta_s  & \vartheta_y
	\end{array} \right]  
	\label{eqn:GT}
\end{equation}

where $\vartheta_s$ is the (in our case) fixed scaling vector. The scaling vector is dependent on the (manually chosen) expected cell size $d_c$. For our data, prior investigation has shown that all typical cells in our case are fully contained within an area of $64\,px$ around the cell center, so with $d_i = 128\,px$ being the length of the input image, we can derive: 
\begin{equation}
	\vartheta_s = \frac{d_i}{d_c} = 0.5
\end{equation} 

The (relative) coordinate grid for the STN is spaced from -1.0 to 1.0, with 0.0 being the center pixel. The translation elements of the ground truth transformation vector in eqn. \ref{eqn:GT} thus become:

\begin{equation}
	\vartheta_{\{x,y\}} = - \frac{2\Delta_{\{x,y\}}}{d_i} 
\end{equation}

\subsection{Network layout}

Our network consists of three main blocks, as depicted in Figure \ref{fig:network}: The localizer, the classifier and the Spatial Transformer Network. The localizer is a deep convolutional network with two stacked convolutional and max-pooling layers, one inception layer and two fully connected layers. It  regresses an estimate $\widehat{\theta}$ of the transform matrix $\theta$.

The classifier is a rather small convolutional neural network with 7 layers, using also convolutional, max-pooling and inception layers. It outputs a vector of dimension 3, which represents the class probabilities for the three cell types depicted in Figure \ref{fig:mitosis}.

Inception \cite{Szegedy:2014tb} blocks were introduced by Szegedy \textit{et al.} in 2014, and have been since then widely used in classification tasks. They are based on the idea that visual information should be processed at different scales, and described to be particularly useful for localization \cite{Szegedy:2014tb}. We incorporated an inception layer, much like Szegedy, between the initial convolutional and max-pooling layers and the fully connected layers. In our case, the inception layer increased convergence and  performance in both localizer and classifier.

\subsection{Training}
The network was trained with the TensorFlow framework using the Adam optimizer. Each image was augmented with an arbitrarily rotated copy of itself to increase robustness of the system. To not assume priors for the cell types, the distributions for the training were made uniform by random deletion of non-minority classes within the training set. A five-fold cross-validation was used.

\subsubsection{Classification network}
In order to achieve good localization and classification performance, we propose a three stage process: In a first step, centered cell images are presented to the network, and the classification-part of the network is trained for 50 epochs using an initial learning rate of $10^{-3}$. This serves as a good initialization of the network for later use. As loss function, denoting the (one hot coded) ground-truth cell class $c$ and the estimated class probabilities $\widehat{c}$, standard cross-entropy is used:

\begin{equation}
    l_\textrm{cla} = -  \sum_{i=1}^3 ln\left(\widehat{c_i}\right) \cdot c_i
\end{equation}

\subsubsection{Training of the localization network}
In the next step, the localization network is trained. For this, the images were cropped with a random offset from the original image, as described in section \ref{chp:methods}. Knowledge of this random offset enables to define a ground truth transformation matrix $\theta$ for optimizing the network. This is used to regress the estimated transformation vector $\widehat{\theta}$ with its elements

\begin{equation}
	\widehat{\theta} = \left[ \begin{array}{ccc}
		\widehat{\vartheta_1} & \widehat{\vartheta_2} & \widehat{\vartheta_x} \\
		\widehat{\vartheta_3} & \widehat{\vartheta_4}  & \widehat{\vartheta_y}
	\end{array} \right]  
	\label{eqn:estimate}
\end{equation}

We want $\widehat{\theta}$ to be an affine transform with no skew and known scale $\vartheta_s$. To achieve this, we first derive the scaling of the estimated transform as:

\begin{equation}
	\widehat{\vartheta_{s_x}} = \sqrt{\widehat{\vartheta_1}^2 + \widehat{\vartheta_3}^2}
\end{equation}
\begin{equation}
	\widehat{\vartheta_{s_y}} = \sqrt{\widehat{\vartheta_2}^2 + \widehat{\vartheta_4}^2}
\end{equation}

Further, we want the diagonal elements $\widehat{\vartheta_1}$ and $\widehat{\vartheta_4}$ to be equal and the off-diagonal elements $\widehat{\vartheta_2}$ and $\widehat{\vartheta_3}$ equal with opposite sign, resulting in a rotation matrix with scale.
These constraints compile into the loss for the localization network: 

\begin{equation}
\begin{split}
l_\textrm{loc} =& 
 \left|\widehat{\vartheta_x} - \vartheta_x  \right|^2 + \left| \widehat{\vartheta_y} - \vartheta_y  \right|^2 + \left| \widehat{\vartheta_{s_x}} - \vartheta_s \right|^2 + \\
 &\left| \widehat{\vartheta_{s_y}} - \vartheta_s \right|^2 + \left| \widehat{\vartheta_1} - \widehat{\vartheta_4} \right|^2 + \left| \widehat{\vartheta_2} + \widehat{\vartheta_3} \right|^2 
\end{split}
\end{equation}

The rotation angle of the transform is a degree of freedom and thus not covered by the loss. The localization part of the network is trained for 200 epochs using an initial learning rate of $10^{-4}$.

\subsubsection{Final refinement of the classification network}
Finally, the whole network is trained for 100 epochs, using an initial learning rate of $10^{-4}$. This final step is calculated on the translated images that were estimated by the localization network and the STN, and it is using a combined loss:  

\begin{equation}
	l = l_\textrm{loc} + \kappa \cdot l_\textrm{cla}
\end{equation}

This loss thus incorporates knowledge about the proper class of the image, about the position of the cell within the image and about the scaling of the patch representing the cell, yet the rotation angle is not known.

\subsection{Baseline comparison}
\label{chp:baseline}
It is hard to compare our results to other authors' works, because unlike them, we consider different cell types within the image and our data set is sparse and not fully annotated. For a baseline comparison, we took a 12-layer CNN like the one described by Cire\c{s}an \textit{et al.} \cite{Ciresan:2013up} 
 for Mitosis detection, but aimed at a three class problem and with an input size of 128x128 px. This classification network was trained for 200 epochs using an initial learning rate of $10^{-3}$.

\section{Results and Discussion}

There were only minor differences in the results of the individual test sets in cross-validation, which is why we concatenated the respective test vectors and calculated the following metrics on the ensemble. We achieved an accuracy of $91.8\,\%$, with precisions reaching from $90.4\,\%$ to $93.4\,\%$ and recall reaching from $90.1\,\%$ to $92.8\,\%$, as described in table \ref{tab:results}. Compared to the baseline CNN described in section \ref{chp:baseline}, this is a significant increase, with the added benefit of retrieving also segmentation information.

\begin{figure}[htb]
  \centering

  \includegraphics[width=.99\linewidth]{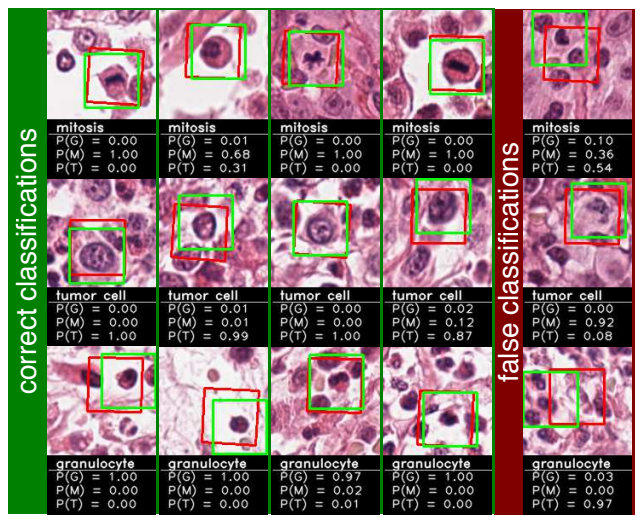}
  \caption{\label{fig:resultsExamples}
           Random choice of correct and false image classifications alongside with selected focus areas, as picked by the localizer. The probabilities denoted are those for the classes: P(G)=granulocytes, P(M)=mitosis, P(T)=normal tumor cells }
\end{figure}

Regarding misclassifications, it is noteworthy that for many false decisions the root cause of error seems to be within the scope of the localizer (see right column of figure \ref{fig:resultsExamples}). In the top and bottom examples depicted there, the localizer selected a different cell than the one originally annotated. Particularly for tumor cells, this is not always a definite fault, since we do not consider annotation information of the direct environment of the annotated cell. If, in a direct surrounding of a tumor cell, a granulocyte or mitotic cell is present, the localizer in fact behaves completely correct in presenting this cell to the classifier. Since we do not aim at finding or classifying all cells, this is no major drawback. In fact, we inherently prioritize classification this way: Since we crop around a known sparse event (mitotic cells or granulocyte), and give this label to our classifier, we incorporate the knowledge that sparse events are more important than others into the loss function.

\begin{table}
\begin{tabular}{|l|l|ccc|}
	
\hline

approach & name 		  &   precision & recall & f1-score  \\
\hline
CNN baseline &
 granulocytes   &   0.847  &   0.898  &   0.872    \\
 &     mitotic figures   &   0.822  &   0.853  &   0.837  \\
        & normal t. cells   &   0.916  &   0.859  &   0.887  \\
 & \textbf{avg / total}   &   \textbf{0.870}  &   \textbf{0.868}  &   \textbf{0.869}  \\
\hline
CNN-STN &
 granulocytes   &   0.912  &   0.925  &   0.918   \\
 &     mitotic figures   &   0.891  &   0.889  &   0.890  \\
        & normal t. cells   &   0.932  &   0.924  &   0.928  \\
 & \textbf{avg / total}   &   \textbf{0.915}  &   \textbf{0.915}  &   \textbf{0.915}  \\
\hline
\end{tabular}
	\caption{Overall classification results of the proposed network. }
	\label{tab:results}
\end{table}

We think that the acquired data set provides a good fundament for further approaches in mitosis detection, where in our opinion the lack of a sufficient amount of samples may limit the methodic progress. 

The acquired data set is also a very interesting candidate for transfer learning. Assuming that many known CNN-approaches suffer from networks that partially do not have well defined filters due to lack of training data, in-domain transfer learning from our mitosis data to other, fully labeled data sets like the competition data sets should be beneficial.

\section{Summary}

In this work the potential of Spatial Transformer Networks within a convolutional neural network approach, applied to segmentation and classification tasks in digital histology images, has been shown. 

The presented approach focuses the attention of a classification network to a part of the original image where most likely a sparsely distributed cell type (mitosis or granulocyte) can be found. 

Further, we have acquired and introduced a data set of cell images from different classes of H\&E stained histology images, with at least ten thousand pathologist-rated samples per class. 

Modeling the localization and classification process independently but with a joint training cuts down on computational complexity of the overall system. 
We believe that this work is an important step towards a microscope-embeddable algorithm that can help the pathologist in counting of mitotic figures by finding a representative area within a histology slide, an algorithm which could reduce inter-rater-variability and thus improve overall quality of tumor grading systems.


\bibliographystyle{eg-alpha-doi}

\newcommand{\etalchar}[1]{$^{#1}$}


\end{document}